\documentclass[default,iicol]{sn-jnl}

\usepackage[flushleft]{threeparttable}

\jyear{2023}%

\theoremstyle{thmstyleone}%
\theoremstyle{thmstyletwo}%

\theoremstyle{thmstylethree}%

\begin{document}

\title[Synthetic Data Generation]{Synthetic Data Generation for Bridging Sim2Real Gap in a Production Environment}


\author*[1]{\fnm{Parth} \sur{Rawal}}\email{parth.rawal@ifam.fraunhofer.de}

\author[1]{\fnm{Mrunal} \sur{Sompura}}
\author[1,2]{\fnm{Wolfgang}~\sur{Hintze}}

\affil[1]{\orgname{Fraunhofer Institute for Manufacturing Technology and Advanced Materials (IFAM)}, \orgaddress{\street{Ottenbecker Damm 12},  \postcode{21684} \city{Stade}, \country{Germany}}}

\affil[2]{\orgdiv{Institute of Production Management and Technology (IPMT)}, \orgname{Hamburg University of Technology TUHH}, \orgaddress{\street{Denickestraße 15},  \postcode{21071} \city{Hamburg}, \country{Germany}}}


\abstract{Synthetic data is being used lately for training deep neural networks in computer vision applications such as object detection, object segmentation and 6D object pose estimation. Domain randomization hereby plays an important role in reducing the simulation to reality gap. However, this generalization might not be effective in specialized domains like a production environment involving complex assemblies. Either the individual parts, trained with synthetic images, are integrated in much larger assemblies making them indistinguishable from their counterparts and result in false positives or are partially occluded just enough to give rise to false negatives. Domain knowledge is vital in these cases and if conceived effectively while generating synthetic data, can show a considerable improvement in bridging the simulation to reality gap. This paper focuses on synthetic data generation procedures for parts and assemblies used in a production environment. The basic procedures for synthetic data generation and their various combinations are evaluated and compared on images captured in a production environment, where results show up to 15\% improvement using combinations of basic procedures. Reducing the simulation to reality gap in this way can aid to utilize the true potential of robot assisted production using artificial intelligence.}




\maketitle

\section{Introduction}\label{sec1}

The revolution in the computer graphic hardware in the last decade has given birth to countless new Artificial Intelligence (AI) applications. Some of these applications are object detection, segmentation as well as pose estimation using deep learning. The time and effort needed for labelling data varies in different applications. While object classification only needs category labels for training, object detection and segmentation need bounding box and pixels coordinates respectively. In case of object pose estimation, scene data such as 6D pose of all the objects in the images is needed which makes annotating real images extremely challenging \cite{Borrego.16.07.2018, Tremblay.18.04.2018, Mayershofer.2021}. Hence, for these applications, generating fast and easy synthetic data with annotations is indeed the preferred way in comparison to the time consuming and manual practice of labelling real data.

As far as training with synthetic data is concerned, a number of successful approaches have been proposed and validated for datasets containing household objects with considerate number of features. However, the components used in production are often textureless, reflective and colorless making them difficult to detect as compared to objects with varying features \cite{C.A.Akar.2022, Moonen.2023}. Moreover, objects trained in a generic environment with domain randomization tend not to work so well for a specialized domain such as production environment \cite{A.Prakash.2019}. For such scenarios, target domain knowledge is extremely essential to further bridge the simulation to reality gap \cite{Mayershofer.2021, Eversberg.2021}. Particularly for the manufacturing industry, there are several challenges to be addressed while working with Convolution Neural Network (CNN). Complex assembly CAD (Computer Aided Design) models contain thousands of parts which must be manually selected and exported as meshes for training. The individual parts after their integration into assemblies are often occluded and go undetected. Another problem is true detection of similar parts with slight differences. In addition to that, hundreds of unseen objects lying inside a production cell, pose a risk of getting detected as false positives.

This paper presents a solution to the previously mentioned problems in production facilities. In the beginning, an approach is proposed and implemented to automatically segregate the target parts from their assembly CAD model and export them in the form of mesh, which is used later for generating data. In case of synthetic data generation, not all of the domain randomization aspects can be combined in a single scene. For example, it has to be decided whether a texture or a background image needs to be chosen for a scene. In these cases, the generated dataset is a combined one without prior knowledge of their individual efficacies in real environments \cite{Tobin.2017, Tremblay.18.04.2018, Nowruzi.16.07.2019}.

\begin{figure}[h]%
\centering
\includegraphics[width=1\columnwidth]{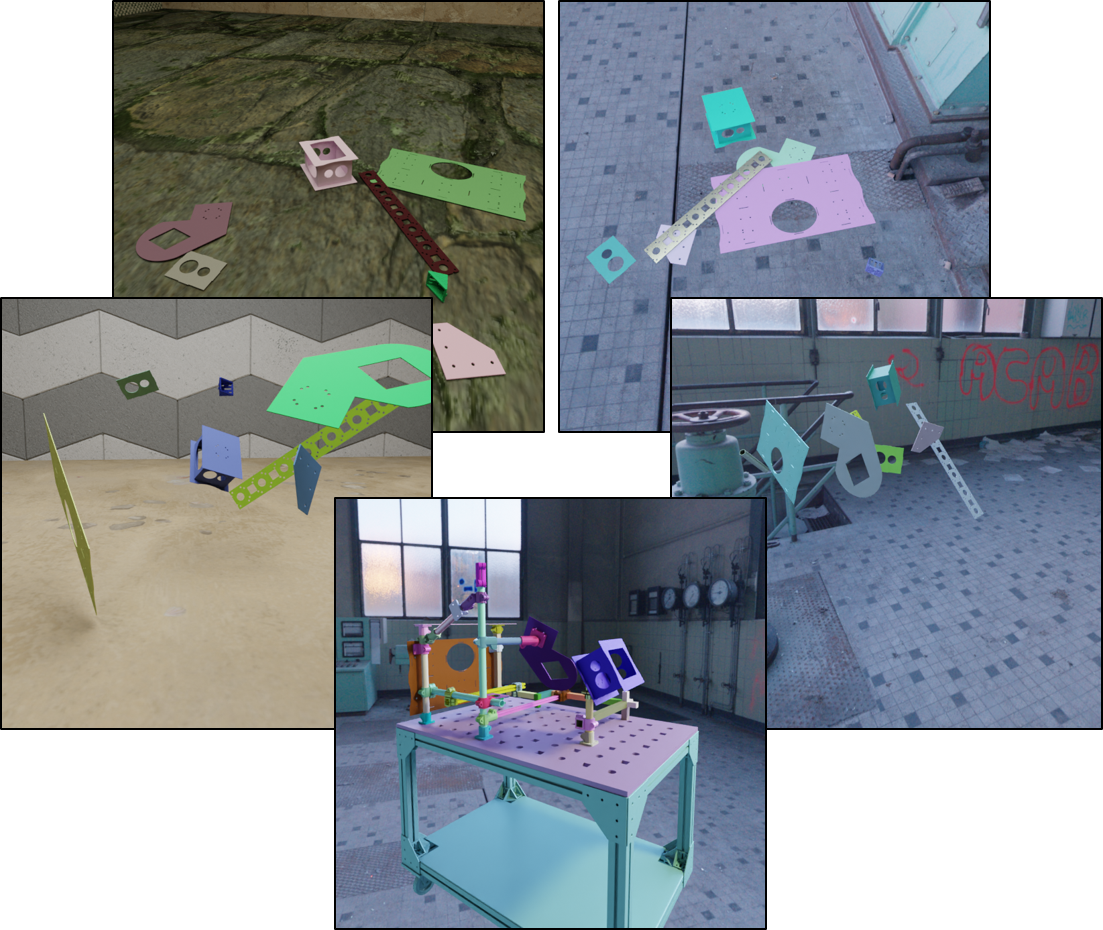}
\caption{Synthetic images generated with five different domain randomization procedures}
\label{fig_recipes}
\end{figure}

In this paper, the focus lies completely on various data generation procedures and photorealistic images are generated with the state of the art data generation tools. Five basic procedures with different levels of domain randomization and domain adaption are presented. Fig.~\ref{fig_recipes} shows images from different basic procedures with different levels of domain randomization. The data generated from the basic procedures is then evaluated on real images inside a production cell. Based on the performance of the basic procedures, synthetic data is generated by combining basic procedures in different proportions and is thereafter evaluated. Some of the combinations boost up the performance by 15\% inside a production cell. The proposed work is developed in the form of an automated pipeline being able to generate thousands of images within few hours with a desired combination of basic procedures. The generated data can be used in robot assisted manufacturing applications for object classification, detection, segmentation and 6D pose estimation tasks.

The following section provides an overview of related work in the area of synthetic data generation and domain randomization and their applications in industrial environments. The reusable procedure based data generation pipeline is described in detail in Section~\ref{sec3}. The evaluation of the data generated by procedures and their combinations on real images captured in a production environment is presented in Section~\ref{sec4}. Finally, the discussion and outlook are mentioned in Section~\ref{sec5}.

\section{Related Work}\label{sec2}

Synthetic data generation is a dominating field in computer vision research in the last five years. Over the years, numerous different ways of generating data are presented. Based on the state-of-the-art methods, these can be broadly divided into four parts. Firstly, cut and paste or render and paste where patches of objects are cropped and pasted on randomly or at specific realistic locations on background images after scaling and transforming them \cite{D.Dwibedi.2017, Georgakis.25.02.2017}. Data can be generated cheaply in this way, but they lack 3D scene information due to which they are not suitable for generating annotations for object pose estimation. Second approach uses physics engines or simulators, whose reviews are compiled by Collins et al. in \cite{Collins.2021}. The physics simulators are primarily used in the field of reinforcement learning due to a vast range of availability of sensors in the test environment. However, Borrego et al. in \cite{Borrego.16.07.2018} and Tobin et al. in \cite{Tobin.2017} point out a number of limitations using MuJoCo and Gazebo for generation of synthetic images. The third approach for generating quick domain randomized data uses game engines like Unreal Engine (UE4) and Unity3D and is depicted by Tremblay et al. \cite{Tremblay.18.04.2018} and To et al. \cite{To.2018}. Eventhough the game engines can render synthetic images reasonably well, they are tuned to provide realtime performance and lack dynamic characteristics of photorealism \cite{Smid.2017, Morrical.28.05.2021}. Finally, the last method consists of physically based rendering (PBR), a technique which can achieve high level of photorealism. Cycles engine from Blender is an example of it. Mayershofer et al. in \cite{Mayershofer.2021} conclude that a network trained on images generated with physics based rendering (Cycles) outperforms the one that was trained on images rendered by a game engine (EEVEE).

Normally, generation of synthetic data, owing to the unavailability of the application programming interface (API), can be a daunting process and is not scalable. However, a number of recently developed pipeline projects on rendering engines have enabled a completely scalable data generation process. NDDS \cite{To.2018} project, based on Unreal Engine (UE4), was used to generate a dataset  in \cite{J.Tremblay.2018}. BlenderProc \cite{Denninger.25.10.2019} procedural pipeline supporting scripts, based on the Cycles engine from Blender, also showed improved performance by their generated data \cite{Denninger.2020}.  A very similar project, NViSII \cite{Morrical.28.05.2021} was demonstrated on NVIDIA's OptiX ray tracing engine. A more recent pipeline Kubric \cite{Greff.07.03.2022} used Blender and PyBullet to generate photorealistic data with ground truth.

The techniques of domain randomization and domain adaption have been widely used to bridge Sim2Real gap \cite{Borrego.16.07.2018, Ren.10.03.2019, S.Hinterstoisser.2019, Tobin.2017, Tremblay.18.04.2018}. While domain randomization is very robust to varying environment, domain adaption helps to achieve a higher precision across domains. These projects show that domain randomization can be easily implemented for synthetic data thereby generalizing the trained model to the real world and improve its performance. Domain adaption, on the other hand, reduces the gap by increasing the resemblance between the two domains. This can be done by either using a combination of synthetic and real images \cite{Y.Huangfu.2021, Baaz.09.12.2022}, or by using synthetic photorealistic images resembling to the target domain \cite{J.Dummel.2021}, or by using generative adversarial networks (GAN) \cite{Goodfellow.2014} based methods which can be used to transform generated synthetic images on the target domain \cite{A.Shrivastava.2017, S.Sankaranarayanan.2018, X.Peng.2018, P.Rojtberg.2020}. However, domain adaption techniques, often being tricky, require greater manual effort compared to domain randomization as mentioned by Borrego et al. in \cite{Borrego.16.07.2018}. The use of photorealism together with domain randomization leads to higher confidence values and enables training models without freezing the backbone layers \cite{T.Hodan.2019, Denninger.2020}. Some of the later projects have shown that combining domain randomization with target domain knowledge improves detection performance \cite{A.Prakash.2019, Mayershofer.2021, Eversberg.2021}.

As far as reducing Sim2Real gap in a target domain is considered, almost all research papers use domain adaption techniques along with domain randomization. These have been presented for different applications in the fields of autonomous driving, logistics and industrial production \cite{F.Reway.2020, Y.Huangfu.2021, Mayershofer.2021, Schoepflin.2022, Dummel.2021, Eversberg.2021, Baaz.09.12.2022}. D{\"u}mmel et al. in \cite{Dummel.2021} use Autodesk Inventor modelling software for data generation and clearly show a performance improvement for an assembly use case. This was achieved by loading the relevant CAD models directly for rendering images of individual parts as well as entire assemblies. However, the data generation software lacks physics simulation and uses Unity3D as a simulation engine to get realistic orientations of the parts. While Moonen et al. \cite{Moonen.2023} and Baaz et al. \cite{Baaz.09.12.2022} generate synthetic images based on rendering pipeline from Unity,  Mayershofer et al. \cite{Mayershofer.2021}, Eversberg and Lambrecht \cite{Eversberg.2021} and Andulkar et al. \cite{M.Andulkar.2018} use Blender API for data generation, making all the mentioned approaches scalable. \cite{Moonen.2023} and \cite{Eversberg.2021} also show improvements for industrial use cases. Out of all the target domain based studies, only two works \cite{Mayershofer.2021, Eversberg.2021} show a comparison between data generated with different parameters and their effect on results. This paper generalizes the data generation methods using an approach of basic domain randomization procedures. Further, it compares the object detection performance for different combination of procedures in a production-related environment. Additionally, the paper also proposes a pipeline to handle complicated CAD models and export the parts and assembly models to mesh format with ease. The final outcome is an overall pipeline to generate synthetic data with a desired combination of procedures. This pipeline being already built on the scalable BlenderProc \cite{Denninger.25.10.2019} pipeline, makes the entire data generation process scalable for multiple applications.

\section{Synthetic Data Generation Pipeline}\label{sec3}
This section introduces the scalable pipeline approach for synthetic data generation in detail and the demonstrator used later for validation of the generated images.
\subsection{Overall Concept}\label{sec31}
One of the attractive features of data generation is to have a reusable pipeline, where desired training data can be generated without additional effort or with minimum effort. For this reason, the current pipeline is built on the top of the scalable BlenderProc \cite{Denninger.25.10.2019} pipeline. A comparable framework, NViSII \cite{Morrical.28.05.2021} can also be used as an alternative to BlenderProc pipeline. The goal here is to generate training data from CAD models without manual efforts, making the framework reusable across multiple industrial applications. The procedures for data generation are implemented within the framework and enable the user to generate desired training data within a few hours. The vision of the framework is to generate the training data only once, while using it for training of multiple AI applications such as object classification, detection, segmentation and pose estimation.

\begin{figure*}[h]
\centering
\includegraphics[width=1\textwidth]{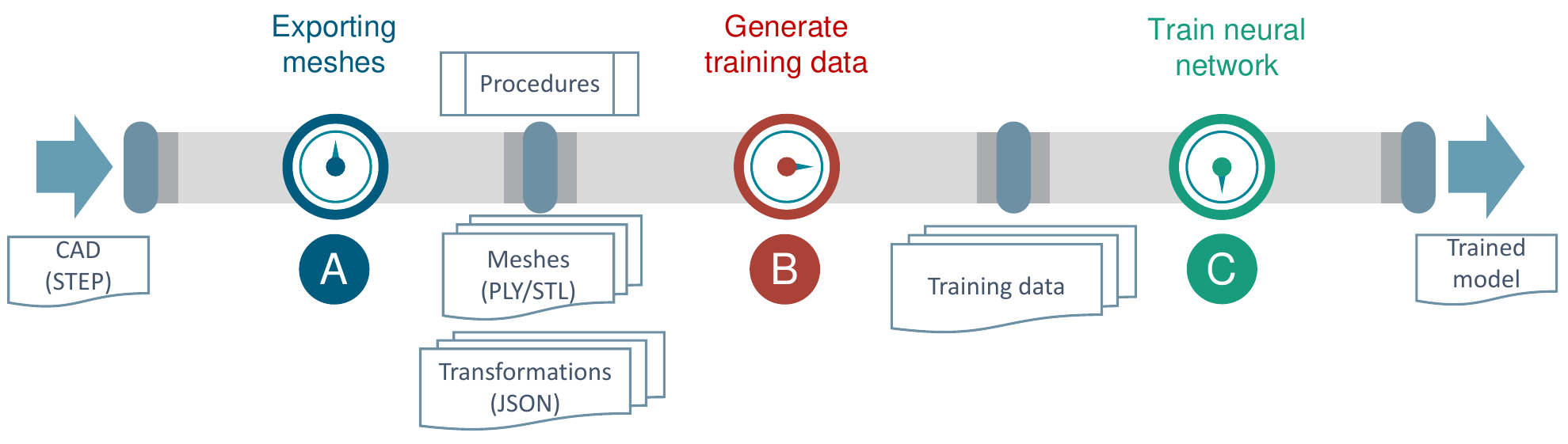}
\caption{Scalable pipeline concept from CAD model to trained model with interfaces inbetween}
\label{fig_pipeline}
\end{figure*}

The structure of the pipeline is depicted in Fig.~\ref{fig_pipeline}. The pipeline can be divided into further three pipelines. Pipeline A reads the CAD model in the form of a STEP file and exports the parts of interest in mesh format for data generation. This is implemented using the free and open-source tool \emph{FreeCAD}\footnote{\href{https://www.freecad.org/}{https://www.freecad.org/}}. Pipeline B takes the mesh data and generates training data with predefined procedures using \emph{BlenderProc}\footnote {\href{https://github.com/DLR-RM/BlenderProc/}{https://github.com/DLR-RM/BlenderProc/}} \cite{Denninger.25.10.2019}. Finally, pipeline C is the neural network training pipeline, which is not within the scope of this paper. Pipeline A and pipeline B are presented in detail in Section~\ref{sec32} and in Section~\ref{sec33} respectively. This pipeline approach encompasses overall data generation process starting from CAD model till training data generation. Additional tools are also developed to aid the visualization of ground truth generated by the data generation pipeline.

In order to validate the generated data in a real production related environment, a special demonstrator constructed for Skotty research project\footnote{Skotty - Smart, collaborative, multitechnology production systems for coating aircraft components, Project timeline: 08.2021-12.2023 under grant ZW 1 80159842 is conducted with support of the Lower Saxony Ministry of Economic Affairs, Employment, Transport and Digitalization and the N-Bank} was used. The demonstrator consists of multiple parts which are simplified versions of components used in aviation industry in case of passenger aircrafts. The final application in Skotty research project is spray coating the components with wax.

\begin{figure}[h]%
\centering
\includegraphics[width=1\columnwidth]{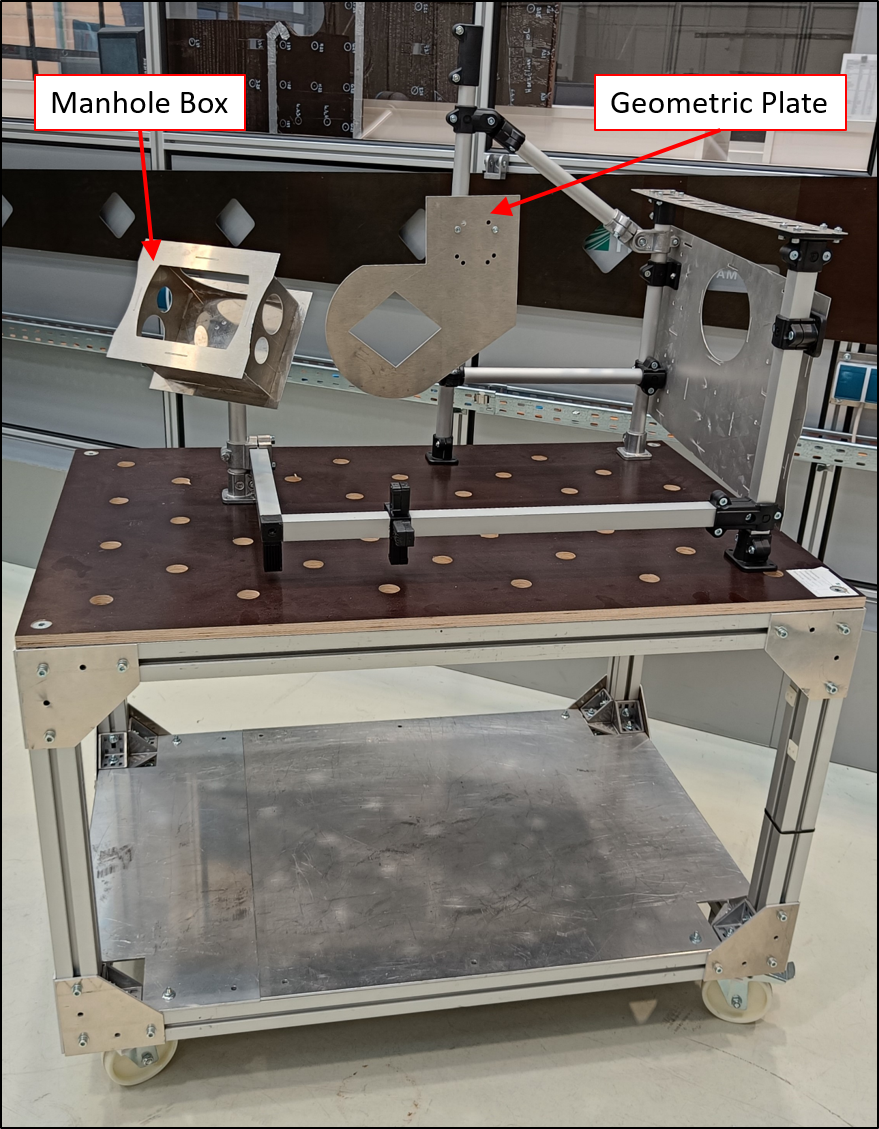}
\caption{Demonstrator used for validation of synthetic data in real environment}
\label{fig_demonstrator}
\end{figure}

Fig.~\ref{fig_demonstrator} shows an image of the demonstrator used along with its various parts and sub-assemblies mounted on an industrial hand cart. The STEP model of the main assembly consists of altogether 148 components, comprising of various parts, sub-assemblies and the main assembly. Out of all components, only \emph{ManholeBox} and \emph{GeometricPlate} are chosen as categories for data generation whereas others are considered as passive objects in the scene. Both the components are asymmetric and contain some features helpful for object detection. While \emph{ManholeBox} has distinctive features on all six sides, \emph{GeometricPlate}, being a planar object, has significant features only on its top and bottom sides. This helps to validate the synthetic data for industrial parts having a planar geometry and their overall effect on the trained model.

\subsection{Exporting Meshes}\label{sec32}

STEP file format is widely used during drafting components due to its accurate geometric representation. In the field of computer graphics, mesh formats like STL and PLY are however more popular, which are approximations of geometric models.  Hence, conversion from STEP format to mesh formats is an essential step for data generation as mesh formats are widely used for photorealistic renderings.

\begin{figure}[h]%
\centering
\includegraphics[width=1\columnwidth]{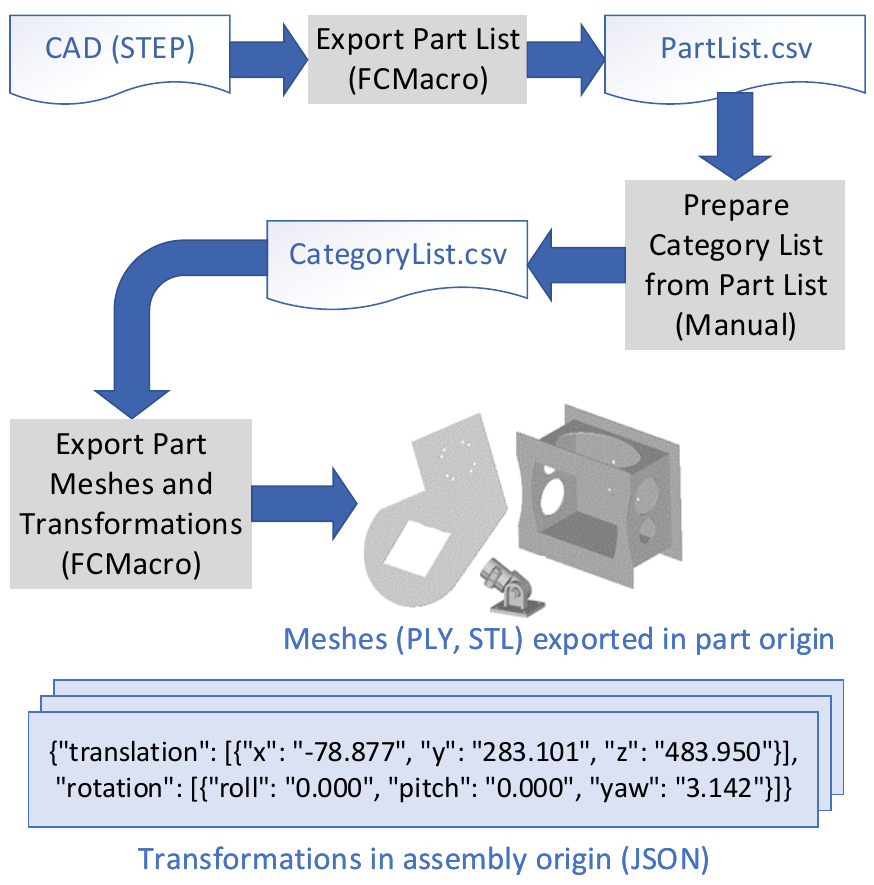}
\caption{Multiple steps for exporting part meshes and their transformations in assembly origin}
\label{fig_exporting_meshes}
\end{figure}

Segregating part files from assembly CAD model containing thousands of parts can be cumbersome and labour intensive. This pipeline partly automates the mesh generation process from assembly CAD model using scripts. For this, the \emph{FreeCAD} open-source tool is used which supports macros. The steps undertaken to export the target mesh files from a CAD model are depicted in Fig.~\ref{fig_exporting_meshes}. At first, the assembly STEP model is loaded in \emph{FreeCAD}. A macro exports all the part, sub-assembly and main assembly file names to a CSV file called here as \emph{PartList.csv}. As the next step, the user manually assigns a category name to all the parts or sub-assemblies to be classified and is saved as \emph{CategoryList.csv}. This step is not extensive as normally not all the components from the assembly CAD model are needed for training. In the final step, another macro automatically generates the mesh files for all the components mentioned in \emph{CategoryList.csv} and places them in a local folder \emph{Classes}. The components which are in \emph{PartList.csv} but not \emph{CategoryList.csv} are considered as passive objects and their meshes are exported in a local folder \emph{Structure}. It is noteworthy that the meshes are exported in the local coordinate frame, in which the parts were drafted and not in the main assembly coordinate frame. In addition, the transformation of the individual components in the main assembly coordinate frame are exported in JSON format as text files in the folders \emph{Classes} and \emph{Structure}. These transformations are a part of target domain knowledge and can be later used to recreate the scene where the components can be placed with positions and orientations as in the main assembly model. Transformations are saved in millimeter and as Euler angles in Roll-Pitch-Yaw convention as shown in Fig.~\ref{fig_exporting_meshes}.

The macro scripts used here are based on Python having file extension \emph{FCMacro} and use \emph{FreeCAD API} to extract useful information from the CAD models. The final outputs are the meshes and transformations in the folders \emph{Classes} and \emph{Structure}. The proposed pipeline is easy to use without much adaption and can export the necessary meshes within a few minutes.

\subsection{Procedure based Data Generation}\label{sec33}
The creation of synthetic datasets involves artificial generation of data to mimic real world scenarios, which in turn requires the use of object meshes, photorealistic textures, real world backgrounds, and other elements for PBR. The resulting images can be either photorealistic or heavily randomized with random choice of colors. In order to establish a pipeline that can be used with different objects or combinations of rendering techniques, various procedures are proposed. These procedures can be imagined to be similar to cooking recipes. Just like the cooked dish may vary based on the cooking recipe followed, the synthetic images generated here also depend on the type of procedure used. These procedures include creating a 3D scene with textured planes or by using a background image with high dynamic range (HDR). Objects are either simulated to fall on the ground plane using physics simulation or are positioned randomly thereby floating in the 3D space. In context of image generation, objects on the ground plane exhibit a deterministic behavior in the ground truth, whereas randomly posed floating objects demonstrate a stochastic behavior. For instance, a circular plate shaped planar object on being dropped onto a surface is unlikely to assume a vertical or angled position after undergoing physics simulation, thereby reducing randomness in the dataset. However, if the same plate is floating in 3D space with a camera perspective such that only the flat edges are seen and not its distinctive features, high losses while training can be observed. Combining these approaches helps to overcome these limitations and reduce Sim2Real gap. Table~\ref{table1} provides a list of all the procedures and their respective variations.

\begin{table}[h]
\begin{center}
\caption{List of procedures for synthetic data generation}
\label{table1}%
\begin{tabular*}{1\columnwidth}{@{\extracolsep{\fill}}p{0.15\columnwidth}p{0.75\columnwidth}@{\extracolsep{\fill}}}
\toprule
Procedures & Variation \\
\midrule
Procedure1 & Textured planes with objects on the ground\\
Procedure2 & Textured planes with objects in 3D space\\
Procedure3 & HDRI background with objects on the ground\\
Procedure4 & HDRI background with objects in 3D space\\
Procedure5 & HDRI background and target scene reconstruction\\
\botrule
\end{tabular*}
\end{center}
\end{table}

\begin{figure*}[!htpb]
\centering
\includegraphics[width=1\textwidth]{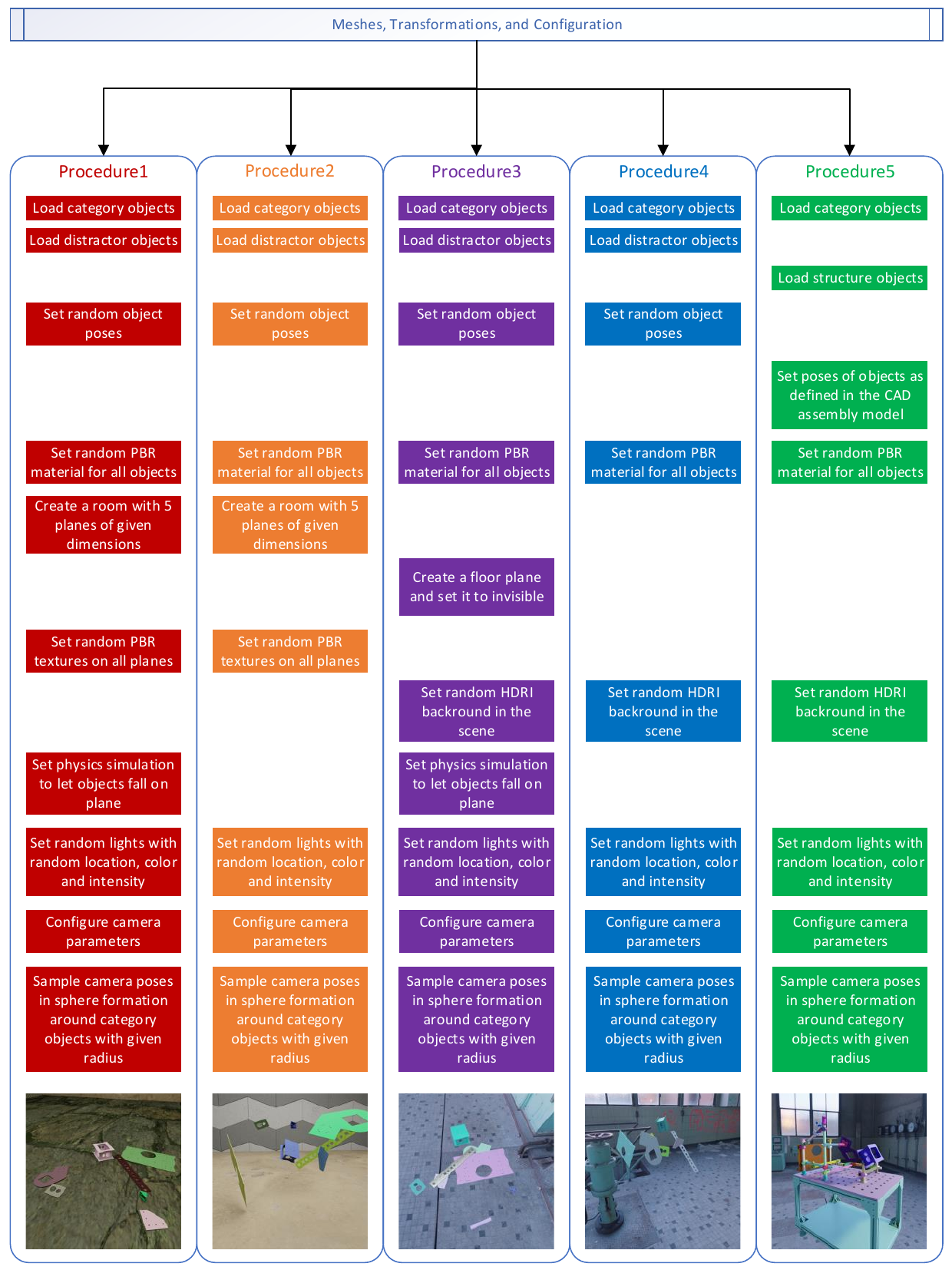}
\caption{Five basic data generation procedures explained in top-down approach}
\label{fig_data_generation_recipes}
\end{figure*}

Fig.~\ref{fig_data_generation_recipes} provides a comprehensive explanation of the step by step data generation process for all procedures in a top-down approach. The configuration file stores various parameters such as the procedure(s) to be used for datasets, camera intrinsic matrix, camera resolution and simulation parameters. To generate the required number of images, each procedure has an outer loop that deals with the number of images required and an inner loop that handles the rendering process, including the addition of objects and distractors to the scene. Object transformations, such as scaling, sampling the pose, etc. are performed for each procedure. Procedure5 differs from all other procedures in terms of setting the object's pose. Specifically in Procedure5, the object's pose is determined using transformations exported to the JSON file in assembly origin, as outlined in Section~\ref{sec32}. On the other hand, for all the remaining procedures, the pose is sampled randomly after the object has been added to the scene. If any objects overlap, they are reset until the condition is satisfied. Procedure1 and Procedure3 use a physics-based approach, allowing the objects to fall down on the ground plane.

Various objects within the scene possess surface appearance properties that are randomized, including but not limited to base colour, roughness, specularity, and metalness. Additionally, the scene incorporates textured planes or backgrounds based on a predetermined procedure. The size of these planes can be customized as per requirement, and they are further enhanced with randomly generated textures sourced from \emph{CC0Textures}\footnote{\href{https://ambientcg.com/}{https://ambientcg.com/}}. In case of procedures with backgrounds, only indoor HDRI backgrounds are used from \emph{PolyHaven}\footnote{\href{https://polyhaven.com/hdris}{https://polyhaven.com/hdris}}. For Procedure3, an invisible plane has been added to the scene as a floor to enable working with physics simulation.

In order to achieve photorealistic results, random lighting conditions are introduced in the rendered scene. Three types of lights are employed: sun, point light source, and light plane. The intensity, colour, and position of each light source are randomized in accordance with the size of the textured plane. In each scene, one of these lights is randomly selected along with random parameters. Once the scene is prepared for image capture, the camera intrinsic matrix is defined. The camera position is sampled on a sphere around the mean location of the objects in the scene. The location of the camera is sampled on the sphere as per the defined radius and the number of image samples required in the inner loop.

\begin{figure*}[!htpb]
\centering
\includegraphics[width=1\textwidth]{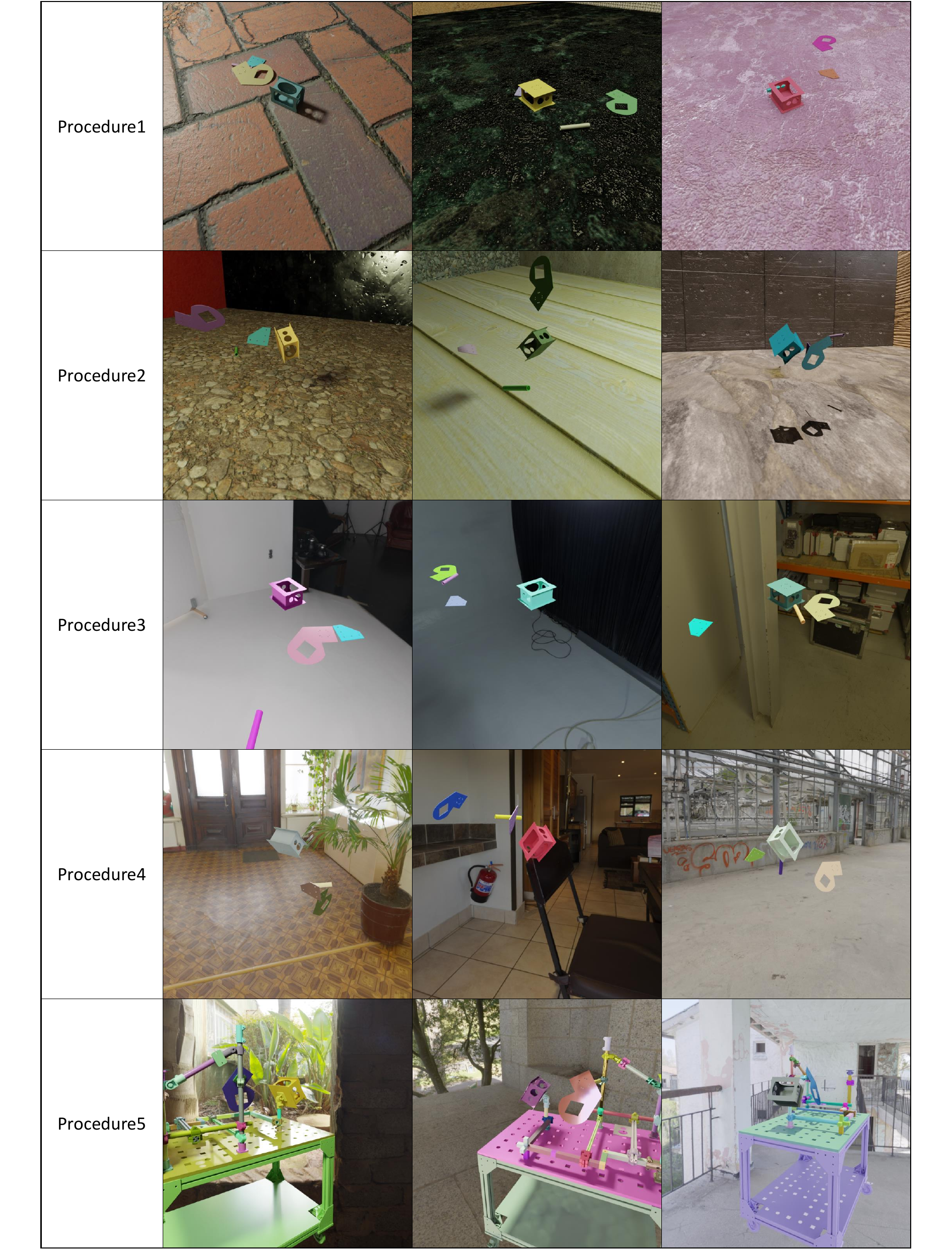}
\caption{Synthetic images generated from basic procedures}
\label{fig_samples}
\end{figure*}

The images based on the sampling of camera locations are rendered in accordance with the defined camera intrinsic matrix. Some samples of these generated images are presented in Fig.~\ref{fig_samples}. The RGB images are saved as arrays within an HDF5 container, along with depth images, class segmentation images, and instance segmentation images that contain information regarding the category ID and name of all objects in the image. This is done using the BlenderProc \cite{Denninger.25.10.2019} pipeline. The use of the HDF5 container is due to its compressibility, extensibility, and simplified I/O operations, which make data storage more manageable. Furthermore, a JSON file in COCO annotation format \cite{Lin.01.05.2014} containing information about the 2D bounding boxes for each category ID, is stored alongside the HDF5 files. In order to utilize the dataset for object pose estimation, 6D object poses are also saved in a JSON file. Finally, camera intrinsic matrix parameters are also stored for future usage. This generated dataset can be utilized in various applications that employ deep learning techniques.

\subsection{Ground Truth Visualization}\label{sec34}
The generated dataset contains a diverse array of images featuring disparate textures and backgrounds. For each image, the pose of an object is stored in the camera coordinate system in the form of homogenous transformation matrix. The translation vector is saved in meters as [X, Y, Z], while the rotation matrix is a 3x3 matrix. This dataset conforms to the BOP \cite{Hodan.24.08.2018} dataset format, which is commonly used for 6D pose estimation. As a result, existing deep learning models can be trained on the generated dataset with only a few minor modifications to the data loader.

\begin{figure}[!htpb]%
\centering
\includegraphics[width=1\columnwidth]{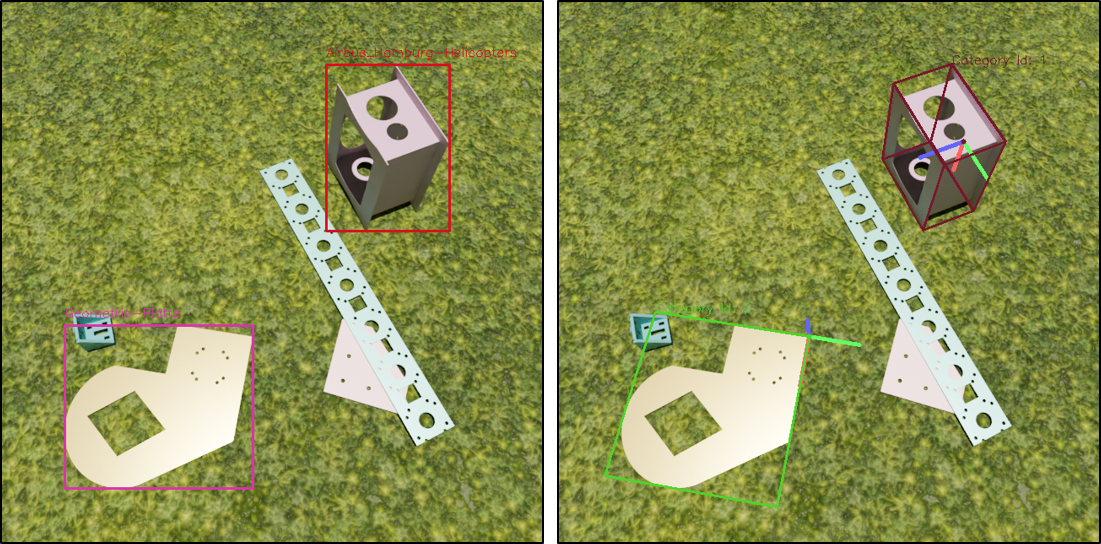}
\caption{Visualization of annotations on images. Left image: Plotting 2D bounding boxes and category labels on image. Right image: Plotting 3D bounding boxes and object coordinate system on image}
\label{fig_visualization}
\end{figure}

In order to verify the ground truth generated along with the synthetic images and for preventing any potential issues arising from category ID discrepancies, the saved ground truth data is visualized where it is superimposed onto the generated RGB images. In Fig.~\ref{fig_visualization}, both 2D and 3D bounding boxes are visualized and plotted on the images for all the present objects. The 2D bounding box information was extracted from the COCO annotations JSON file for each object in each image. For the 3D bounding boxes with origin, object mesh files in PLY format were loaded, and their dimensions were saved in separate JSON files containing information such as diameter, minimum dimensions, and size in the X, Y, and Z directions. Once the position, orientation, and 3D bounding box of an object are obtained, a transformation is applied from the camera coordinate system to the image coordinate system using the camera intrinsic matrix and the 3D bounding box is projected on the image as seen in the right image of Fig.~\ref{fig_visualization}.

\section{Validation}\label{sec4}

The datasets generated with all five basic procedures and their combinations were evaluated for object detection using YOLOv7 \cite{Wang.06.07.2022}. YOLOv7 implementation, with its evaluation scripts, enables quick and easy calculation of object detection metrics. For a fair comparison, all the configuration parameters and hyperparameters were kept unchanged. The normal YOLOv7 model mentioned in \cite{Wang.06.07.2022} was chosen and was trained for 100 epochs. Datasets with 15\,000 images and 70/30 train test split ratio were generated for both the objects together. For combination datasets with a mixed procedures, the same proportion which was used while dataset generation, was maintained for train and test datasets. The time taken to generate a single dataset on a computer equipped with NVIDIA RTX A4000 took between 9 to 12 hours depending on the procedure(s) used. Altogether, five basic procedures and five combination sets of basic procedures were validated on 500 synthetic and 150 real images.

The synthetic images for validation were generated separately with each of the five basic procedures, while the real images were captured with a camera in a production environment and annotated later. The captured real images were taken in two scenarious: first, where both the objects were loose and second, where they were assembled in the demonstrator. For all the scenarios, COCO's standard metric \cite{Lin.01.05.2014} denoted as mAP@[0.5, 0.95] and PASCAL VOC's metric \cite{Everingham.2010} denoted as mAP@0.5 are calculated. However, only mAP@[0.5,0.95] results are presented here as they put a larger emphasis on the localization of bounding boxes \cite{S.Bell.2017}.

\subsection{Evaluation of Basic Procedures}\label{sec41}

Table~\ref{table2} and Table~\ref{table3} draw a performance comparison between basic procedures for \emph{ManholeBox} and \emph{GeometricPlate} respectively. Here, P is the short form for representing the procedures. It can be seen that all the diagonal elements of the 5x5 matrix have the best results across different synthetic validation sets. This behavior is expected as models trained on synthetic data using any single procedure have best results if it is validated on the synthetic dataset created with the same procedure. However, some procedures also perform better on validation datasets generated by other procedures. As an example, the average simulation values for P2 and P4 show that they perform overall better on all procedures.

\begin{table}
\begin{center}
\begin{minipage}{1\columnwidth}
\begin{threeparttable}
\caption{Performance comparison of basic procedures for \emph{ManholeBox}}
\label{table2}
\begin{tabular*}{1\columnwidth}{@{\extracolsep{\fill}}cccccc@{\extracolsep{\fill}}}
\toprule%
\begin{tabular}[x]{@{}c@{}}\emph{ManholeBox}\\mAP@[0.5:0.95]\end{tabular} & \multicolumn{5}{@{}c@{}}{Trained on}\\ \cmidrule(r){1-1} \cmidrule(l){2-6}
Validated on & P1 & P2 & P3 &  P4 & P5 \\
\midrule
P1  & 0.98 & 0.99 &  0.98 & 0.99 & 0.48\\
P2  & 0.75 & 0.99 &  0.83 & 0.98 & 0.53\\
P3  & 0.98 & 0.99 &  0.99 & 0.99 & 0.40\\
P4  & 0.75 & 0.98 &  0.89 & 0.99 & 0.56\\
P5  & 0.43 & 0.91 &  0.69 & 0.90 & 0.99\\
\midrule
Average Sim  & 0.78 & \textbf{0.97} &  0.88 & \textbf{0.97} & 0.59\\
\midrule
Real parts loose  & 0.25 & 0.70 &  0.35 & 0.46 & 0.31\\
Real assembly  & 0.00 & 0.51 &  0.00 & 0.58 & 0.74\\
\midrule
Average Real  & 0.12 & \textbf{0.61} &  0.17 & 0.52 & 0.53\\
\botrule
\end{tabular*}
\begin{tablenotes}
    \item Note: Values in bold indicate the maximum average value across different trained models.
\end{tablenotes}
\end{threeparttable}
\end{minipage}
\end{center}
\end{table}

\begin{table}[h]
\begin{center}
\begin{minipage}{1\columnwidth}
\begin{threeparttable}
\caption{Performance comparison of basic procedures for \emph{GeometricPlate}}
\label{table3}
\begin{tabular*}{1\columnwidth}{@{\extracolsep{\fill}}cccccc@{\extracolsep{\fill}}}
\toprule%
\begin{tabular}[x]{@{}c@{}}\emph{GeometricPlate}\\mAP@[0.5:0.95]\end{tabular} & \multicolumn{5}{@{}c@{}}{Trained on}\\ \cmidrule(r){1-1} \cmidrule(l){2-6}
Validated on & P1 & P2 & P3 &  P4 & P5 \\
\midrule
P1  & 0.96 & 0.98 &  0.97 & 0.98 & 0.19\\
P2  & 0.52 & 0.94 &  0.51 & 0.92 & 0.11\\
P3  & 0.96 & 0.99 &  0.99 & 0.99 & 0.16\\
P4  & 0.58 & 0.95 &  0.60 & 0.96 & 0.07\\
P5  & 0.37 & 0.73 &  0.35 & 0.66 & 0.99\\
\midrule
Average Sim  & 0.68 & \textbf{0.92} &  0.68 & 0.90 & 0.30\\
\midrule
Real parts loose  & 0.70 & 0.03 &  0.34 & 0.52 & 0.00\\
Real assembly  & 0.72 & 0.70 &  0.30 & 0.86 & 0.24\\
\midrule
Average Real  & \textbf{0.71} & 0.36 &  0.32 & 0.69 & 0.12\\
\botrule
\end{tabular*}
\begin{tablenotes}
    \item Note: Values in bold indicate the maximum average value across different trained models.
\end{tablenotes}
\end{threeparttable}
\end{minipage}
\end{center}
\end{table}

The performance of the basic procedures on real dataset drops to a considerable amount which is due to simulation to reality gap. In case of \emph{ManholeBox}, model trained on P2 performs the best on real datasets overall. P2 consists of object images with textured planes without simulating physics or letting the objects float in the space. In contrast to that, for \emph{GeometricPlate}, model trained on P1 performs the best. P1 consists of object images with textured plate with physics simulation, meaning that the objects are lying on the ground plane.

This behaviour too can be explained from the geometry of the objects. \emph{ManholeBox} having geometry on all six sides can be detected easily when placed with random 3D poses in space. On the contrary, \emph{GeometricPlate} being a planar object, has geometrical features only on two of the six sides. A random sampling of poses in space will not work the best as the features are not always visible from all perspectives. In this case, when the object falls on the ground plane, most of the features are always visible to the camera as the object has less probability to fall in a position which places itself into an unstable state. Letting objects fall on the ground plane resemble to a deterministic behavior whereas the objects lying in 3D space resemble to a stochastic behavior. The results show that for planar objects like \emph{GeometricPlate}, synthetic images taken after letting them fall on the ground plane yield better results. For other objects, random sampling of poses as shown in P2 is helpful. Model trained on P4, which is rendering with a random background in 3D space, also shows good results for both objects. This is because backgrounds, even though not photorealistic, do contain high variance of features as compared to textures which are photorealistic.

\begin{table}[h]
\begin{center}
\caption{Composition of combination datasets from basic procedures}
\label{table4}%
\begin{tabular*}{1\columnwidth}{@{\extracolsep{\fill}}cccccc@{\extracolsep{\fill}}}
\toprule
& \multicolumn{5}{@{}c@{}}{Basic procedures}\\ \cmidrule(r){1-1} \cmidrule(l){2-6}
Combinations & P1 & P2 & P3 &  P4 & P5 \\
\midrule
C1  & 20\% & 20\% & 10\% & 30\% & 20\%  \\
C2  & 40\% & 0\% & 0\% & 40\% & 20\%  \\
C3  & 0\% & 40\% & 0\% & 40\% & 20\%  \\
C4  & 0\% & 0\% & 0\% & 80\% & 20\%  \\
C5  & 50\% & 0\% & 0\% & 50\% & 0\%  \\
\botrule
\end{tabular*}
\end{center}
\end{table}

\subsection{Evaluation of Combination Datasets}\label{sec42}

Based on the performance of the basic procedures on the real datasets, five different combination datasets made up of different proportion of synthetic data from basic procedures were generated. Table~\ref{table4} shows the composition of different combination datasets represented from C1 till C5. These datasets were used for training and were validated on simulated and real images. 

\begin{table}[h]
\begin{center}
\begin{minipage}{1\columnwidth}
\begin{threeparttable}
\caption{Performance comparison of combination datasets for \emph{ManholeBox}}
\label{table5}
\begin{tabular*}{1\columnwidth}{@{\extracolsep{\fill}}cccccc@{\extracolsep{\fill}}}
\toprule%
\begin{tabular}[x]{@{}c@{}}\emph{ManholeBox}\\mAP@[0.5:0.95]\end{tabular} & \multicolumn{5}{@{}c@{}}{Trained on}\\ \cmidrule(r){1-1} \cmidrule(l){2-6}
Validated on & C1 & C2 & C3 &  C4 & C5 \\
\midrule
P1  & 0.99 & 0.99 &  0.99 & 0.99 & 0.98\\
P2  & 0.98 & 0.98 &  0.98 & 0.98 & 0.96\\
P3  & 0.99 & 0.99 &  0.99 & 0.99 & 0.99\\
P4  & 0.99 & 0.98 &  0.99 & 0.99 & 0.97\\
P5  & 0.99 & 0.99 &  1.00 & 0.99 & 0.87\\
\midrule
Average Sim  & \textbf{0.99} & \textbf{0.99} & \textbf{0.99} & \textbf{0.99} & 0.95\\
\midrule
Real parts loose  & 0.51 & 0.61 &  0.52 & 0.47 & 0.41\\
Real assembly  & 0.27 & 0.78 &  0.72 & 0.85 & 0.20\\
\midrule
Average Real  & 0.39 & \textbf{0.69}\tnote{*} &  0.62\tnote{*} & 0.66\tnote{*} & 0.31\\
\botrule
\end{tabular*}
\begin{tablenotes}
    \item Note: Values in bold indicate the maximum average value across different trained models.
    \item[*] Better results than the best performing basic procedure.
\end{tablenotes}
\end{threeparttable}
\end{minipage}
\end{center}
\end{table}

\begin{table}
\begin{center}
\begin{minipage}{1\columnwidth}
\begin{threeparttable}
\caption{Performance comparison of combination datasets for \emph{GeometricPlate}}
\label{table6}
\begin{tabular*}{1\columnwidth}{@{\extracolsep{\fill}}cccccc@{\extracolsep{\fill}}}
\toprule%
\begin{tabular}[x]{@{}c@{}}\emph{GeometricPlate}\\mAP@[0.5:0.95]\end{tabular} & \multicolumn{5}{@{}c@{}}{Trained on}\\ \cmidrule(r){1-1} \cmidrule(l){2-6}
Validated on & C1 & C2 & C3 &  C4 & C5 \\
\midrule
P1  & 0.98 & 0.99 &  0.98 & 0.98 & 0.97\\
P2  & 0.94 & 0.90 &  0.94 & 0.91 & 0.83\\
P3  & 0.99 & 0.99 &  0.99 & 0.99 & 0.97\\
P4  & 0.96 & 0.95 &  0.96 & 0.96 & 0.88\\
P5  & 0.99 & 0.99 &  0.99 & 0.99 & 0.54\\
\midrule
Average Sim  & \textbf{0.97} & 0.96 &  \textbf{0.97} & \textbf{0.97} & 0.84\\
\midrule
Real parts loose  & 0.06 & 0.36 &  0.18 & 0.49 & 0.68\\
Real assembly  & 0.86 & 0.94 &  0.82 & 0.96 & 0.81\\
\midrule
Average Real & 0.46 & 0.65 &  0.50 & 0.73\tnote{*} & \textbf{0.74}\tnote{*}\\
\botrule
\end{tabular*}
\begin{tablenotes}
    \item Note: Values in bold indicate the maximum average value across different trained models.
    \item[*] Better results than the best performing basic procedure.
\end{tablenotes}
\end{threeparttable}
\end{minipage}
\end{center}
\end{table}

\begin{table*}
\begin{center}
\begin{minipage}{1\textwidth}
\begin{threeparttable}
\caption{Results summary}
\label{table7}
\begin{tabular*}{1\textwidth}{@{\extracolsep{\fill}}ccccccccccc@{\extracolsep{\fill}}}
\toprule%
\begin{tabular}[x]{@{}c@{}}Both objects\\mAP@[0.5:0.95]\end{tabular} & \multicolumn{10}{@{}c@{}}{Trained on} \\ \cmidrule(r){1-1} \cmidrule(l){2-11}
Validated on & P1 & P2 & P3 &  P4 & P5 & C1 & C2 & C3 &  C4 & C5 \\
\midrule
P1  & 0.97 & 0.99 & 0.98 & 0.99 & 0.34 & 0.99 & 0.99 & 0.99 & 0.99 & 0.98\\
P2  & 0.64 & 0.97 & 0.67 & 0.95 & 0.32 & 0.96 & 0.94 & 0.96 & 0.95 & 0.90\\
P3  & 0.97 & 0.99 & 0.99 & 0.99 & 0.28 & 0.99 & 0.99 & 0.99 & 0.99 & 0.98\\
P4  & 0.66 & 0.97 & 0.74 & 0.98 & 0.31 & 0.97 & 0.96 & 0.97 & 0.97 & 0.92\\
P5  & 0.40 & 0.82 & 0.52 & 0.78 & 0.99 & 0.99 & 0.99 & 0.99 & 0.99 & 0.70\\
\midrule
Average Sim  & 0.73 & 0.95 & 0.78 & 0.94 & 0.45 & \textbf{0.98} & 0.97 & \textbf{0.98} & \textbf{0.98} & 0.90\\
\midrule
Real parts loose  & 0.47 & 0.36 & 0.34 & 0.49 & 0.15 & 0.28 & 0.48 &  0.35 & 0.48 & 0.54\\
Real assembly  & 0.36 & 0.60 & 0.15 & 0.72 & 0.49 & 0.56 & 0.86 & 0.77 & 0.90 & 0.51\\
\midrule
Average Real & 0.42 & 0.48 & 0.25 & 0.60 & 0.32 & 0.42 & 0.67\tnote{*} & 0.56 & \textbf{0.69}\tnote{*} & 0.53\\
\botrule
\end{tabular*}
\begin{tablenotes}
    \item Note: Values in bold indicate the maximum average value across different trained models.
    \item[*] Better results than the best performing basic procedure.
\end{tablenotes}
\end{threeparttable}
\end{minipage}
\end{center}
\end{table*}

Table~\ref{table5} and Table~\ref{table6} draw a performance comparison between combination datasets for \emph{ManholeBox} and \emph{GeometricPlate} respectively. It can be easily seen that in general the combination datasets perform way better for synthetic images generated from all procedure unlike in case of basic procedure mentioned in Section~\ref{sec41}. For real images in a production environment, C2, C3 and C4 perform better than the best performing basic procedure for \emph{Manholebox}, while C4 and C5 perform better than the best performing basic procedure for \emph{GeometricPlate}.

Summarizing the results of models trained on combination datasets, in case of \emph{Manholebox}, there is upto 13\% improvement whereas for \emph{GeometricPlate}, performance improvement of about 5\% is observed. A proper mixture of procedures can yield better performance thereby bridging the simulation to reality gap.

\subsection{Final results and discussion}\label{sec43}

The results from basics and combination datasets for both objects are summarized in Table~\ref{table7}. These values are the mean values of the performance of \emph{ManholeBox} and \emph{GeometricPlate} on validation datasets.

Combination datasets show varied overall performance. While the C2 and C4 combinations show improved performance of 11\% and 15\% respectively with respect to the best performing basic procedure, other combinations do not prove to be effective enough. Referring to Table~\ref{table4} again, C2 is composed of three basic procedures: objects on the ground plane with textures, objects in 3D space with indoor background and objects placed as defined in assembly CAD model with indoor backgrounds. C4 is only composed of two basic procedures: objects in 3D space with indoor background and objects placed as defined in assembly CAD model with indoor backgrounds. Textured planes introduce photorealistic characteristics such as reflection and shadows to the rendered images whereas indoor backgrounds introduce versatility in the images reducing the detected false positives. The standalone performance of the target domain based P5 using assembly CAD and indoor backgrounds is not satisfactory on real images but when this procedure is combined in given proportions with other procedures, a performance boost is observed. However, the best performance yielded by C4 does not consist of photorealistic images with textured planes and implies that the need of photorealistic data using textured planes, even after delivering good results, is not a must for performance improvement.

\begin{figure*}[!htpb]
\centering
\includegraphics[width=1\textwidth]{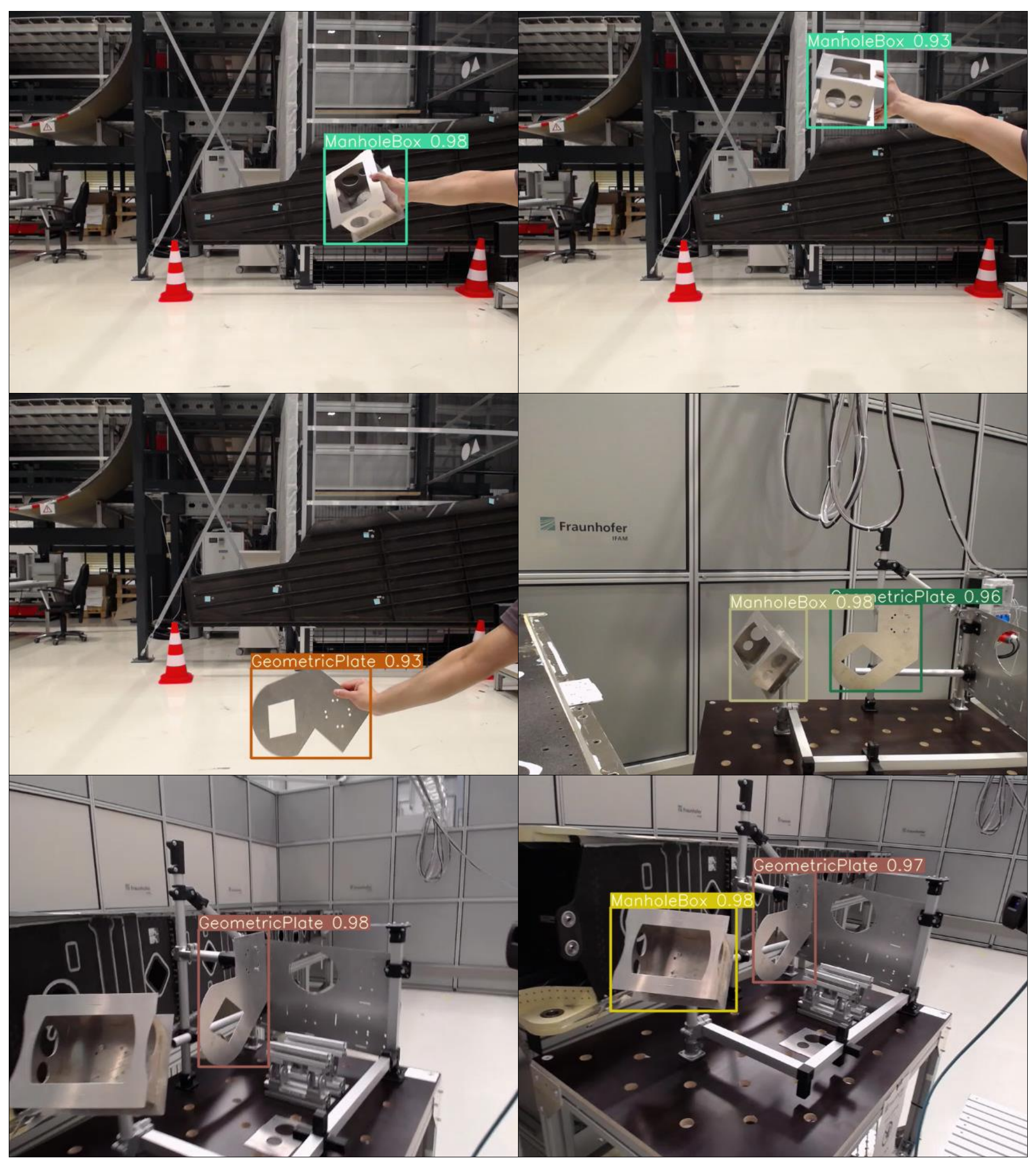}
\caption{Object detection results for both objects with model trained on C2 dataset in a production environment with different scenarios}
\label{fig_results}
\end{figure*}

\begin{figure}[h]%
\centering
\includegraphics[width=1\columnwidth]{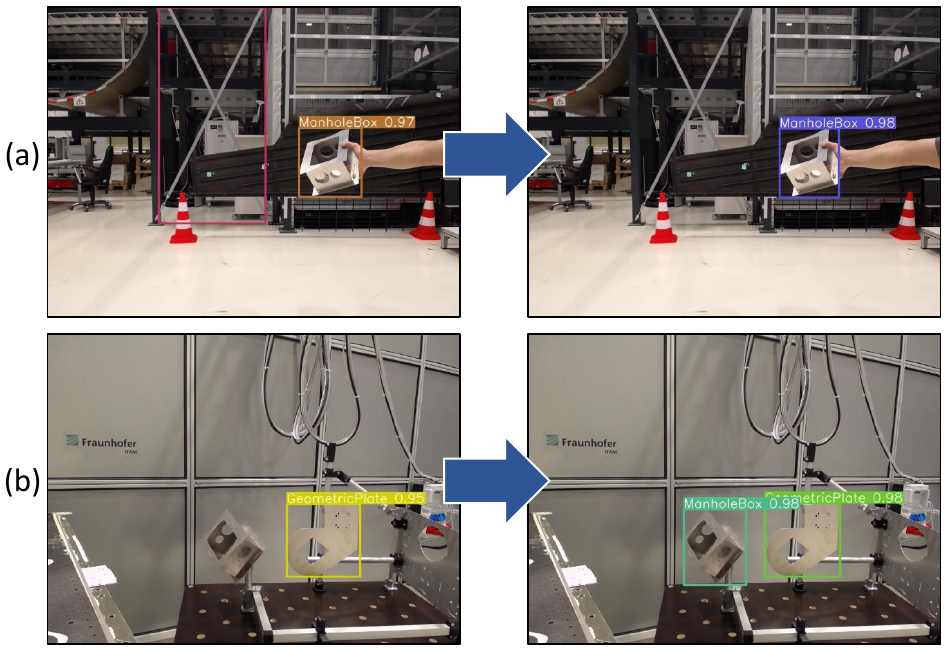}
\caption{Validation images for model trained on C2 dataset showing improved precision and recall over model trained on basic procedure P4. Left images in (a) and (b) show a false positive and a false negative respectively which was correctly handled in the right images using model trained on C2}
\label{fig_improvement}
\end{figure}

Fig.~\ref{fig_results} shows some real world object detection results in a production environment using the trained model on C2 dataset. A score threshold of 0.80 was used while inferencing. Fig.~\ref{fig_improvement} depicts examples of improved performance on combination dataset C2 compared to a basic procedure P4. A reduction in the occurrence of false positives and false negetives can be observed here. For the current example, model trained on C2 dataset was compared with model trained on P4.

\section{Conclusion}\label{sec5}

In this work, a scalable pipeline based approach with procedures based on domain randomization and domain adaption techniques for photorealistic synthetic data generation is presented. Object detection models trained on five basic procedures are validated on real world images inside a production environment and the results are used to derive combination datasets from the best performing procedures. The models thus trained on some of the combination datasets show an improved performance than those which were purely trained on datasets with basic procedures. The final results bridge the Sim2Real gap with about 15\% of performance improvement using combination of basic procedures on real images as compared to the best performing basic procedure. The achieved drastic performance improvement is due to the knowledge of the target domain undertaken while generating synthetic data. This is done by constructing a scene similar to the structure in the CAD model. The models trained with only synthetic images on YOLOv7 \cite{Wang.06.07.2022} show promising performance in a real world production environment.

The pipeline approach presented here starts from generating the mesh files and their transformations in assembly origin from the assembly STEP model using scripts. Then next module is built on the top of BlenderProc pipeline \cite{Denninger.25.10.2019} and generates a synthetic dataset with a desired combination of basic procedures, which can then be used for training neural networks for tasks such that object detection and object pose estimation. This enables easy generation of thousands of annotated images in few hours.

From the validation of the dataset, it can be concluded that inclusion of target domain plays an important role for bridging Sim2Real gap. The use of photorealistic images leads to large improvements and hence reduces the size of training datasets and time required for training. For the future work, some of the GAN based approaches can be tested and evaluated in combination with these procedures. Moreover, the generated synthetic data can be validated for object pose estimation neural networks to check if the same procedures also yield better performance in those cases.

\backmatter

\bmhead{Acknowledgments}

We are grateful for the valuable expert information provided especially by every member of the project team Integrated Production Systems as well as the Head of Stade Branch Dr. Dirk Niermann at Fraunhofer IFAM.

\end{document}